\documentclass[11pt]{article}
\usepackage{tabularx}
\usepackage{times}
\usepackage{url}
\usepackage{latexsym}
\usepackage{relsize}
\usepackage{wrapfig}
\usepackage{ragged2e}
\usepackage{bookmark}
\usepackage{coling2020}
\usepackage{arabtex}
\usepackage{arydshln}
\usepackage{lipsum}  
\usepackage{enumitem}
\usepackage{amssymb}
\usepackage{amsmath}
\usepackage{graphicx}
\usepackage{utf8}
\usepackage[table]{xcolor}
\usepackage{booktabs}
\usepackage{adjustbox}
\usepackage{float}
\usepackage{diagbox}
\usepackage{footnote}
\usepackage{verbatim}
\usepackage[bottom]{footmisc}
\usepackage{tablefootnote}
\usepackage{tabularx}
\usepackage[toc,page]{appendix}
\usepackage{stackengine}
\usepackage{multirow}

\hypersetup{
    colorlinks=true,
    linkcolor=blue,
    filecolor=blue,  
    citecolor=blue,
    urlcolor=blue,
}
\usepackage{comment}
\usepackage{textcomp}
\usepackage{ulem} 
\usepackage[show]{boxnotes}
\newcommand{\eat}[1]{} 

\def\textsubscript#1{\ensuremath{_{\mbox{\textscale{.6}{#1}}}}}

\colingfinalcopy 

\title{Machine Generation and Detection of Arabic Manipulated and Fake News} 

\author{El Moatez Billah Nagoudi$^1$, AbdelRahim Elmadany$^1$, Muhammad Abdul-Mageed$^1$, \\ \textbf{Tariq Alhindi$^2$, Hasan Cavusoglu$^3$} \\

$^1$ Natural Language Processing Lab,\\
$^{1,3}$ The University of British Columbia \\ 
  $^2$ Department of Computer Science, Columbia University  \\
     \small{$^1$ \{moatez.nagoudi,a.elmadany,muhammad.mageed\}@ubc.ca,} \\
    \small{ $^2$ {tariq@cs.columbia.edu}, $^3$ {cavusoglu@sauder.ubc.ca} }
     }

\date{}
\begin{document}
\setcode{utf8}
\maketitle

\begin{abstract}
\normalsize
\noindent Fake news and deceptive machine-generated text are serious problems threatening modern societies, including in the Arab world. This motivates work on detecting false and manipulated stories online. However, a bottleneck for this research is lack of sufficient data to train detection models. We present a novel method for automatically generating Arabic manipulated (and potentially fake) news stories. Our method is simple and only depends on availability of true stories, which are abundant online, and a part of speech tagger (POS). To facilitate future work, we dispense with both of these requirements altogether by providing AraNews, a novel and large POS-tagged news dataset that can be used off-the-shelf. Using stories generated based on AraNews, we carry out a human annotation study that casts light on the effects of machine manipulation on text veracity. The study also measures human ability to detect Arabic machine manipulated text generated by our method. Finally, we develop the first models for detecting manipulated Arabic news and achieve state-of-the-art results on Arabic fake news detection (macro $F_{1}=70.06$). Our models and data are publicly available. \\
\end{abstract}


\section{Introduction}


\blfootnote{
    
     \hspace{-0.65cm}  
     This work is licensed under a Creative Commons 
     Attribution 4.0 International License.
     License details:
     \url{http://creativecommons.org/licenses/by/4.0/}.
}

\begin{wrapfigure}{R}{0.54\textwidth}
\includegraphics[scale=0.39]{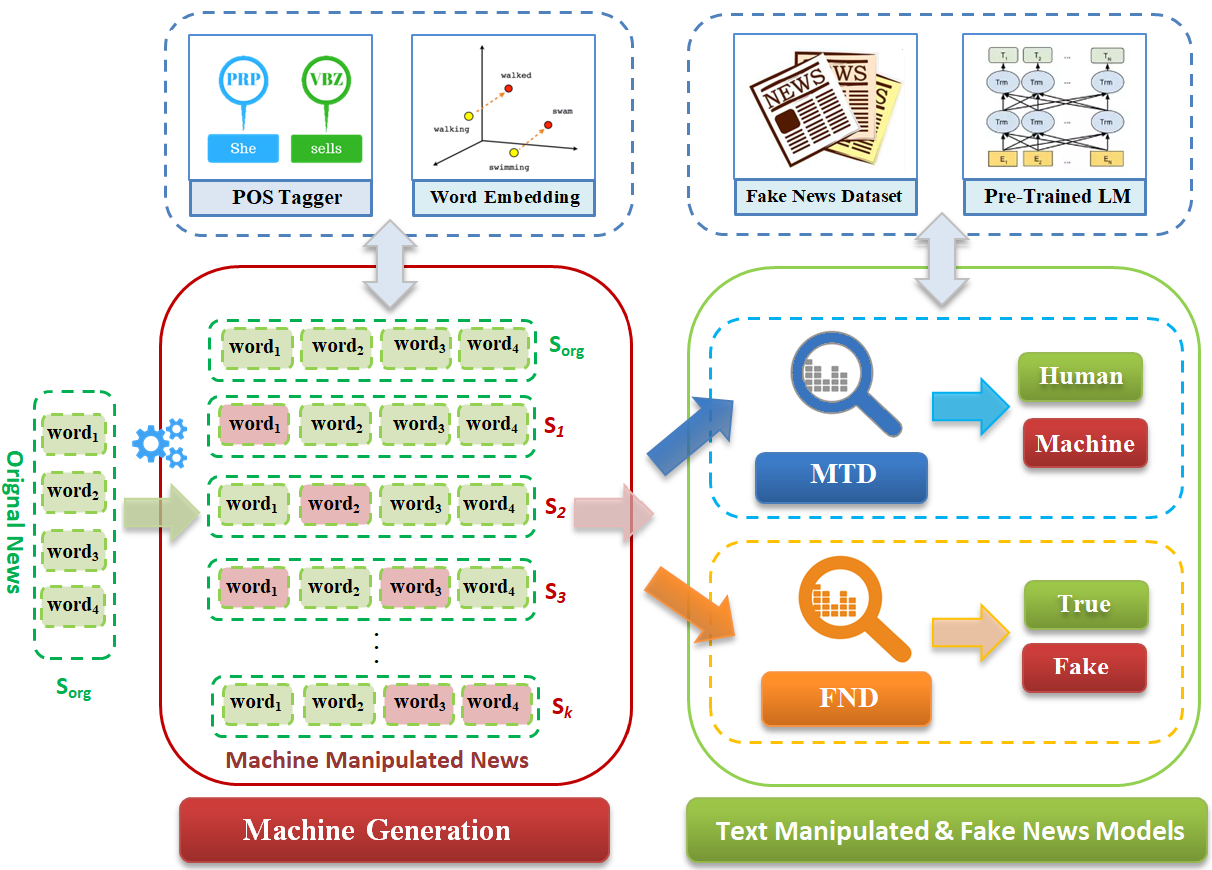}
  \caption{Our proposed methods. \textbf{Left:} Machine generation of manipulated text. \textbf{Top Right:} manipulated text detection model (MTD). \textbf{Bottom Right:} fake news detection model (FND). \textbf{ }\colorbox{green!25}{\textbf{word$_i$}}: original word. \colorbox{red!20}{\textbf{word$_j$}}: substituted word. }
  \label{fig:news_map}
\end{wrapfigure}

The last few years witnessed a striking rise in creation and dissemination of fake news~\cite{egelhofer2019fake,allcott2019trends}. Such fake stories are propagated not only by individuals, but also by groups or even nation states~\cite{allcott2019trends}. For example,~\newcite{allcott2017social} discuss the role fake news have played in the 2016 U.S. presidential election, arguing that Donald Trump's voters have been more influenced to believe fake stories. More recently, concerns have also been raised about possible abuse of machine-generated text such as by GPT3~\cite{brown2020language} for deceiving readers.  

In the Arab context, Arab countries have had their share of misinformation. This is especially the case due to the sweeping waves of uprisings and popular protests ~\cite{torres2018epistemology,helwe2019assessing}. Although there has been considerable research investigating the legitimacy, or lack thereof, of news in many languages \cite{conroy2015automatic,kim2018leveraging,bondielli2019survey}, work on the Arabic language is still lagging behind. 

In this paper, we first report an approach to automatically generate manipulated (and possibly fake) stories in Arabic. Our approach is simple: Given a dataset of legitimate news, a part-of-speech (POS) tagger, and a word embedding model, we are able to automatically generate significant amounts of news stories. Since these generated stories are machine manipulated such that original words (e.g., named entities, factual information such as numbers and time stamps) are substituted, some of these stories can be used as training data for fakes news detection models. 

To illustrate our method, we provide the following scenario: Given a human-authored sentence, we output a manipulated version of the original. The veracity of the  manipulated version can either: (1)~\textbf{Stay~Intact.} For instance when changing an adjective with its synonym, e.g.,  \footnotesize{\<أفضل>}  \normalsize ~(``top") with \footnotesize{\<أحسن>} \normalsize ~(``best") in \footnotesize{\<أفضل هاتف ذكي هو الأيفون>} \normalsize (``The best smartphone is the iPhone")  or (2) \textbf{Change.} For example, when substituting a named entity with another that does not necessarily communicate the meaning of the original as closely. For example, changing the named entity \footnotesize \<أرامكو> \normalsize (``Aramco") with \footnotesize \<أمازون> \normalsize (``Amazon")  in  \footnotesize \<أرباح> 	\<أعلى> 	\<السنة> 	\<هذه> 	\<تحقق> 	\<أرامكو> \normalsize (``Aramco achieved the highest profit this year").

As such, we emphasize that changing a certain POS does not automatically flip the sentence veracity. For example changing \footnotesize{\<مصر>}  \normalsize (``Egypt") with \footnotesize{\<المحروسة>} \normalsize (``Almahrousa") does not alter the sentence veracity. We manually validate the claim that our method of text manipulation can generate fake stories via a human annotation study (Section~\ref{sec:annot}). We then use our generated data to create models that can detect manipulated stories from our method and empirically show the impact of exploiting our generated stories on the fake news detection task on a manually-crafted external dataset (Section~\ref{sec:dete}). We make our models and data publicly available.\footnote{Models and data are at: \url{https://github.com/UBC-NLP/wanlp2020_arabic_fake_news_detection}. } 

We make the following contributions: (1) We introduce AraNews, a new large-scale POS-tagged news dataset covering a wide range of topics from diverse sources. (2) We propose a simple, yet effective, method for automatic manipulation of Arabic news texts. Applying this methods on AraNews, we create and release the first dataset of manipulated Arabic news dataset to accelerate future research. (3) We perform a human annotation study to measure the ability of native speakers of Arabic to detect (a) machine manipulated and (b) fake news stories without resorting to external resources such as fact checking websites. The annotation study aims at gauging the extent to which a human can fall prey to deceptive news in a semi-real situation (i.e., where an average reader do not check third party sources when reading through a news story). (4) We develop effective models for detecting manipulated news stories, and then test the utility of our generated data for improving fake news detection on an external dataset. 



    
    
    

The rest of the paper is organized as follows: Section~\ref{sec:RW}  provides an overview of related work.  In Section~\ref{sec:data}, we describe the two \textit{true}\footnote{We use the terms ``true" and ``legitimate" interchangeably to refer to stories that are not ``fake".} news datasets used in this work. Section~\ref{sec:autogen} is about our methods for generating manipulated text (and potentially fake news stories). Section~\ref{sec:annot} describes our human annotation study. In Section~\ref{sec:dete}, we present our detection models. We conclude in Section~\ref{sec:conc}.

\section{Related Work}
\label{sec:RW}

{\bf Knowledge-Based Fact Checking.} Recent work on developing automatic methods for fake news detection has mainly followed two lines of research as categorized in the literature~\cite{thorne2018automated,potthast2018stylometric}. First, work that compares a claim against an evidence from (trusted) collections of factual information whether the evidence is a sentence (i.e. fact-checking modeled as textual entailment) or a full document (i.e. stance detection between a claim-document pair). This includes work that created synthetic claims verified against Wikipedia~\cite{thorne2018fever}, and naturally occurring claims verified against news articles \cite{ferreira2016emergent,pomerleau2017fake}, discussion forums~\cite{joty2018joint}, or debate websites \cite{chen2019seeing}. These datasets are labeled using $2$ tags (\textit{true}, \textit{false}) \cite{alhindi2018evidence} $3$ tags (\textit{supported}, \textit{refuted}, \textit{not-enough-information}) \cite{thorne2018fever}, or $4$ tags (\textit{agree}, \textit{disagree}, \textit{discuss}, \textit{unrelated})~\cite{pomerleau2017fake}. They vary in size from $300$ claims~\cite{ferreira2016emergent} to $185,000$ claims~\cite{thorne2018fever}. Approaches on developing models to predict claim veracity using these datasets include hierarchical attention networks~\cite{ma2019sentence}, pointer networks~\cite{hidey2020deseption}, graph-based reasoning~\cite{zhou2019gear,zhong2019reasoning}, and (similar to our methods) fine-tuning of pre-trained transformers~\cite{hidey2020deseption,zhong2019reasoning}.  

{\bf Style-Based Detection.} The second line of research focuses on analyzing the linguistic features of a claim to determine its veracity without considering external factual information. This approach is based on investigating linguistic characteristics of fake content in comparison to true content. In news and various fact-checked political claims,~\newcite{rashkin2017truth} found that first and second person pronouns, superlatives, modal adverbs, and hedging are more prevalent in fake content, while concrete and comparative figures, and assertive words are more widespread in truthful content. Other work found the properties of deceptive language to differ between domains~\cite{perez2018automatic}. Misleading content itself has been classified into sub-categories such as (a) the $3$ types of fake (serious fabrication, hoaxes, and satire)~\cite{rubin2015deception}, (b) propaganda and its different techniques~\cite{da2019fine}, and (c) misinformation and disinformation~\cite{ireton2018journalism}. The differences between these different categories depend on many factors such as genre and domain, targeted audience, and deceptive intent~\cite{rubin2015deception,rashkin2017truth}. In addition to categories, truth was classified to more than two \textit{levels}. For example, \texttt{Politifact.com} introduced $6$ levels: pants-on-fire, false, mostly-false, half-true, mostly-true and true. These different levels have been exploited in previous work, with a goal to automate this more challenging six-way classification task~\cite{rashkin2017truth,wang2017liar,alhindi2018evidence}. 

{\bf Automatic Generation of Data.} The development of automatic fake news detection models was possible as the afore-mentioned datasets became available. More related to our work, previous work has focused on developing methods to automatically generate more robust, and large-scale, fake news datasets.~\newcite{thorne2019evaluating} showed that current fact-checking systems are vulnerable to adversarial attacks by doing simple alteration to the training data. To increase robustness of such systems, previous work has extended available fake news datasets both manually and automatically using lexical substitution~\cite{alzantot2018generating}, rule-based alterations~\cite{ribeiro2018semantically}, phrasal addition and temporal reasoning~\cite{hidey2020deseption}, or using transformer models such as GPT-2~\cite{radford2019language} and Grover~\cite{zellers2019defending} for claim and news article generation~\cite{niewinski2019gem,zellers2019defending}.  As a way to increase our understanding and trust in fact-checking systems,~\newcite{atanasova2020generating} developed a transformer-based model for generating fact-checking textual explanations along with the prediction of claim veracity.  

{\bf Arabic Work.} All of the datasets described above, however, are in English with limited availability of similar ones in other languages such as Arabic. Available Arabic datasets cover tasks such as determining claim check-worthiness of tweets~\cite{barron2020checkthat}, news and claims from fact-checking websites~\cite{elsayed2019overview}, and translated political claims from English~\cite{nakov2018overview}. In addition, there are datasets for stance and factuality prediction of claims from news or social media with or without the evidence retrieval task ~\cite{baly2018integrating,khouja2020stance,elsayed2019overview,alkhair2019arabic,darwish2017improved}. These corpora are created by either using credibility of publishers as proxy for veracity (\textit{true}/\textit{false}) then manually annotating the stance between a claim-document pair (\textit{agree}, \textit{disagree}, \textit{discuss}, \textit{unrelated})~\cite{baly2018integrating} or by manual alteration of true claims to generate fake ones about the same topic~\cite{khouja2020stance}--all requiring a manual, slow, and labor-intensive process. We alleviate this by introducing our simple and scalable approach for automatic generation of Arabic manipulated text, including potential fake stories, using the abundant legitimate online news data as seeds for the generation model. We also introduce a large-scale dataset in true and manipulated form for detection work. We now introduce our datasets.

\section{Datasets} 
\label{sec:data}
\subsection{ATB: Arabic TreeBank} \label{sec:ATB}


We exploit a number of Arabic Treebank datasets from the Linguistic Data Consortium (LDC). Namely, we use $4$ LDC resources comprising Arabic news stories in Modern Standard Arabic (MSA). These are: Arabic Treebank (ATB) Part $1$ v$4.1$ (LDC2010T13), Part $2$ v$3.1$ (LDC2011T09), Part $3$ v$3.2$ (LDC2010T08) and Broadcast News v$1.0$ (LDC2012T07),  the latter being a collection of Arabic news stories built as part of   of the DARPA TIDES project.\footnote{\url{https://www.ldc.upenn.edu/collaborations/past-projects}.} These 4 parts contain over $2,000$ news stories produced by a handful of Arabic news services with a total of $1.5$M tokens.  Moreover,  we use the Arabic Treebank  Weblog (LDC2016T02), which contains $13$K Arabic  news and a total of $308$K tokens. We refer to all the 5 LDC resources collectively as \textbf{\texttt{ATB}}. For each token in ATB, there is a Latin-based transliteration, a unique identifier (lemma ID), a breakdown of the constituent morphemes (prefixes, stem, and suffixes), POS tag(s), and the corresponding English gloss(es).

\subsection{AraNews: A New Large-Scale Arabic News Dataset}\label{sec:AraNews}
\begin{wrapfigure}{R}{0.50\textwidth}
  \centering
\begin{centering}
  \frame{\includegraphics[scale=0.16]{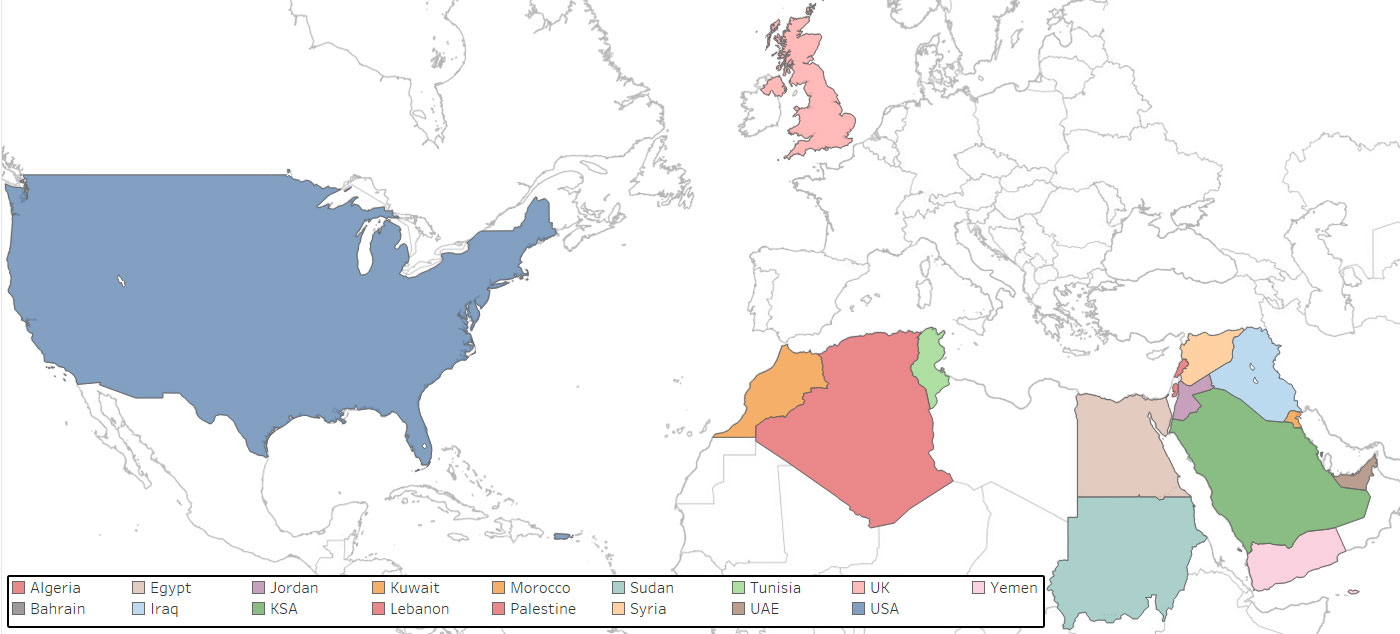}}
  \caption{\small Geographical distribution of AraNews.} \label{fig:img-geo}  \end{centering}
\end{wrapfigure}
In order to study misinformation in Arabic news, we develop, \textbf{\texttt{AraNews}}, a large-scale, multi-topic, and  multi-country Arabic news dataset. To create the dataset, we start by manually collecting a list of $50$ newspapers belonging to $15$ Arab countries, the United States of America (USA), and the United Kingdom (UK). Then, we  scrape the news articles from this list of newspapers. Ultimately, we collected a total of  $5,187,957$ news articles. The map in Figure~\ref{fig:img-geo} shows the geographic distribution of AraNews.

\noindent We assign each article in AraNews a thematic category as follows: We first consider the category assigned on each newspaper website to the article. We identify a total of $118$ unique categories, which we manually map to only $17$ categories using the dictionary illustrated in Table~\ref{tab:norm} in Appendix~\ref{App:norma}. The $17$  categories are in the set \textit{\{Politics, History, Society,  Media,  Entertainments, Weather, Sports, Social Media, Heath, Culture and Art , Economy, Religion, Education, Technology, Fashion,  Local News, International News\}}. For each article in the AraNews collection, we document several types of information. These include: (1) name of the newspaper in Arabic and English, (2) newspaper origin country, (3) newspaper link,  (4)  title, (5)  content, (6)  summary (if available), (7)  author (if available), (8)  URL, (9) date, and (10) topic. More details about AraNews are in Table~\ref{tab:araNews} in Appendix~\ref{subsec:appendix-AraNews-stat}. AraNews is available for research.\footnote{\label{note1}\url{https://github.com/UBC-NLP/wanlp2020_arabic_fake_news_detection}.}


 
%



\section{Methods}\label{sec:autogen}
To generate a large scale manipulated  news dataset, we exploit ATB (see Section~\ref{sec:ATB}) and  $1$M news articles extracted from AraNews (Section~\ref{sec:AraNews}). In the following, we describe our data splits and methodology for automatically generating manipulated text from these two `legitimate' sources.\footnote{We do not check the veracity of stories in these two sources, but we have no reason to think they may have fake stories. As such, we make the assumption they consist of ``true" stories.}

\subsection{Data Splits}
We split both ATB and AraNews at the article level into TRAIN, DEV, and TEST. Table~\ref{tab:data_splits} provides the related statistics at both the article and sentence levels across the different data sources for all three splits.

 \begin{table}[h]
 \begin{center}
 \footnotesize
 \begin{tabular}{lccccccccccc}
       \toprule
\multicolumn{1}{l}{\small \textbf{Data}}                & \multicolumn{3}{c}{\small \textbf{TRAIN} (80\%)}                                                & \multicolumn{3}{c}{\small  \textbf{DEV}  (10\%)}                        & \multicolumn{3}{c}{\small  \textbf{TEST} (10\%)}                     &   \\   \cline{2-10}  
 \multicolumn{1}{l}{}                    & \multicolumn{1}{l}{\small  Artic. } & \multicolumn{1}{l}{\small  Sent.} & \multicolumn{1}{l}{\small  Tokens}& 
 \multicolumn{1}{l}{\small  Artic.} & \multicolumn{1}{l}{\small  Sent.} &\multicolumn{1}{l}{\small  Tokens}  
 &\multicolumn{1}{l}{\small  Artic. } & \multicolumn{1}{l}{\small  Sent.}&\multicolumn{1}{l}{\small  Tokens} \\ \toprule  
\textbf{\small ATB Weblog }   & $1.9$K                  & $8.6$K    & $154.8$K   & $235$               & $1$K    & $17.9$K    & $235$                & $1.2$K    & $19.5$K    \\
\textbf{\small ATB Part 1 }  &  $587$                    & $4.7$K    & $117.3$K    & $73$                & $580$   & $14.9$K      & $74$                & 536   & $13.3$K       \\
\textbf{\small ATB Part 2  } & $400$                    & $3.4$K    & $117.6$K    & $50$                & $387$     & $13.9$K     & $51$                & 382    & $12.7$K      \\ 
\textbf{\small ATB Part 3}   & $479$                    & $10.5$K  & $268.5$K     & $60$                & $1.5$K     & $36.5$K   & $60$                & $1.4$K   & $34.7$K     \\ 
\textbf{\small ATB BN} & $96$                     & $21.5$K   & $334.3$K   & $12$                & $3.1$K     & $47.7$K  & $12$                & $2.4$K   & $40$K \\   

\hline  

\hline
\textbf{\small AraNews}               & $800$K                  & 	$3.3$M   & $1.1$B     & $100$K               & $55.1$K     & $209.9$M   & $100$K               &  $61.6$K  &  $197.2$M  \\

 \toprule
 \end{tabular}
     \caption{\small Statistics of ATB and AraNews (only 1M articles) datasets across the data splits.}
    \label{tab:data_splits}
    \end{center}

 \end{table}

\subsection{POS Tagging}
The first step in our approach is to perform POS tagging of the news articles. ATB is already POS tagged. Thus, we use MADAMIRA~\cite{pasha2014madamira}, a morphological analysis and disambiguation tool for Arabic, to POS-tag AraNews.\footnote{MADAMIRA was trained on the training sets of  Penn Arabic Treebank corpus (parts 1, 2 and 3)~\cite{maamouri2004} and   the  Egyptian  Arabic  Treebanks~\cite{maamouri-etal-2014-developing}.}

\subsection{News Word Embedding Model}  The second  component needed in our model is a word vector model. We train a fastText model~\cite{joulin2016fasttext} on a concatenation of MSA data sources (Wikipedia Arabic,\footnote{\url{https://archive.org/details/arwiki-20190201}.} Arabic Gigaword Corpus~\cite{parker2009arabic},  and ATBP1V3~\footnote{\url{https://catalog.ldc.upenn.edu/LDC2010T08}}). 
We perform light pre-processing involving removing punctuation  marks, non-letters, URLs, emojis, and emoticons. We also convert elongated words back to their original form by reducing consecutive repetitions of the same character as suggested in~\cite{lachraf-etal-2019-arbengvec}. For example : \footnotesize{\<استفسااااار>} \normalsize (\textit{inquiries}) \normalsize and \footnotesize{\<الجزائــــــــــر>} \normalsize(\textit{Algeria}) are converted to \footnotesize{\<استفسار> and \<الجزائر>}. \normalsize We then train our model using the Python Gensim library~\cite{vrehuuvrek2011gensim}. We set  the vector size to $300$, minimum word frequency at $100$, and a window size of $5$ words.  We call this model \textbf{\texttt{AraNewsEmb}}. We then use this model to retrieve the most similar tokens of a given token in the original text using cosine  similarity. 
 Next, we use one of the set of relevant tokens to replace the original token,  focusing only on tokens corresponding to the following POS tags: proper nouns (\small{N\_PROP}\normalsize), cardinal numbers (\small{N\_NUM}\normalsize), common adjective (\small{ADJ}\normalsize), comparative adjective (\small{ADJ\_COMP}\normalsize), ordinal numbers (\small{ADJ\_NUM}\normalsize), and negative particles (\small{NEG\_PART}\normalsize). In theory, substitution of these words should have no syntactically harmful effect on the sentence. However, changes can happen if the gold or predicted POS tag is wrong.

\subsection{Automatic Text Manipulation} 
\begin{wraptable}{r}{0.38\textwidth}
 \centering
 \footnotesize

 \begin{tabular}{lrcc}
 \toprule
POS Label & Count  & Avg  & Median \\
\toprule
ADJ        & $99,538$ & $3.79$   & $3.00$   \\
\footnotesize ADJ\_COMP  & $4,513$  & $2.81$   & $2.00$   \\
\footnotesize ADJ\_NUM   & $5,752$  & $3.06$   & $3.00$   \\
\footnotesize N\_NUM  & $60,615$ & $0.55$   &  $0.00$   \\
\footnotesize sqN\_PROP & $75,771$ & $2.93$   & $2.00$  \\
\toprule 
\end{tabular}
\caption{\footnotesize Descriptive statistics of $k-$closest words excluded in each POS class. We simply remove the negation token corresponding to \textbf{NEG\_PART} from the sentence, and so the embedding model is not used in this case.}  \label{tab:pos_exc}

\end{wraptable}
\normalsize To generate a machine manipulated story, we substitute the selected words (ones matching the listed POS tags) by a chosen one from the $k$ most similar ($k$-closest) words in our AraNewsEmb model as described in~\cite{nagoudi-schwab-2017-semantic}. We remove negation from the sentence, using the negative particle (\small{NEG\_PART}\normalsize) POS as a guide, and substitute the cardinal number related to (\small{N\_NUM}\normalsize) with a random number. For tokens related to the rest of POS tags, we needed to identify a reasonable \textit{character-level} similarity threshold between the original token and the retrieved most-similar token to ensure the two belong to different lemmas.~\footnote{We use the following formula to compute the character-level similarity ratio between two tokens: $ratio=2*M/T$, where $M$ is matching characters and $T$ is total of characters.}

\noindent We performed a manual analysis based on $5,000$ random substitution examples from AraNewsEmb and identify a similarity $ratio$ of $50\%$. This threshold gave us new words in $100\%$ of the cases. For instance,  if we want to substitute the word   \footnotesize{\<لبنان>} \normalsize (Lebanon),  we exclude three words:  \footnotesize{\<ولبنان>, 	\<لبنانيا>, 	\<بلبنان>, } \normalsize  before considering the $4^{th}$-closest word which is  \footnotesize{\<سوريا> } \normalsize (Syria). Other examples for the substitution process are illustrated in  Table~\ref{tab:k-close}. We also provide in Table~\ref{tab:pos_exc} the average number of $k$-closest words excluded in each POS class. The results of this step are two new machine manipulated datasets. We refer to these datasets as \textbf{\texttt{ATB$^\textbf{+}$}}  and \textbf{\texttt{AraNews$^\textbf{+}$}}. More details about these two datasets are in Table~\ref{tab:man-data} in Appendix~\ref{App:data-man}. We now provide an example illustrating how our text manipulation method works.

 \begin{table}[ht]
\centering
\begin{adjustbox}{width=15cm}
\footnotesize
\renewcommand{\arraystretch}{1}
{

\begin{tabular}{rlllc }
\toprule
\small \textbf{Word} & \textbf{\small Translation} &   \textbf{\small POS}&    \textbf{ \small $k$-closet  (ratio similarity\%)}  & \textbf{\small Token rank}\\
\toprule

\colorbox{white!10}{\<قصير>} & Short &  ADJ   & 	\small{\colorbox{green!15}{\<طويل>}	(25\%) }&  0\\

\colorbox{white!10}{{\<أكثر>}} &{More} &{ADJ\_COMP}   & \small{\colorbox{red!15}{\<واكثر>} (89\%),		\colorbox{green!15}{\<أقل>}	(28\%) }&  1\\

\colorbox{white!10}{{\<باكستان>}} &{Pakistan} &{N\_PROP}   &  \colorbox{red!15}{\<لباكستان>} (93\%),	\small{\colorbox{red!15}{\uline{\<اوزباكستان>}} (82\%),	\colorbox{green!15}{\<بنغلاديش>}	(26\%)} & $2$ \\

\colorbox{white!10}{{\<الثالث>}} &{The third} &{ADJ\_NUM}   & 	\small{\colorbox{red!15}{\<والثالث>}	(92\%),\colorbox{red!15}{\uline{\<الثاني>}} (67\%), \colorbox{red!15}{\uline{\<الاول>}}(72\%)  , \colorbox{green!15}{\<الرابع>}	(49\%)}& 4\\

\colorbox{white!10}{{\<وسبعة>}} &{And seven} &{N\_NUM}   & 	\small{\colorbox{red!15}{\<وسبع>}	(89\%),\colorbox{red!15}{{\<وسبعه>}} (80\%), \colorbox{red!15}{\uline{\<وسبعون>}}(73\%)  , \colorbox{green!15}{\<وثلاثون>}	(17\%)}& 4\\

\toprule
\end{tabular}}
\end{adjustbox}
\caption{\small Illustration of substitution process based on the word embeddings model. \textbf{ Token rank:} refers to rank of chosen word in the returned word embedding list (from AraNewsEmb) after applying our char-based cosine similarity threshold. \colorbox{red!15}{\textbf{Light red:}} excluded word. \colorbox{green!15}{\textbf{Light green:}} selected word. \textbf{\uline{Under lined}} words represent the false negative of the selection process (i.e., words based on a different lemma and hence could work but were ignored by the algorithm).}  \label{tab:k-close}
\vspace{-5mm}
    \end{table}

\subsection{Illustrative Example}

We present a typical example illustrating the automatic text manipulation  process by our method.  Consider the following sports news sentence: \noindent \<دولار> 	\<مليون> 	120	\<مقابل> 	\<برشلونة> 	\<الى> 	\<ينتقل> 	\<محرز>  ~(``Mahrez moves to Barcelona for \$ $120$ million"). The method proceeds in the following steps:
\paragraph{Step 1: Identify POS tags.} The sentence can be POS-tagged as shown in Table~\ref{tab:pos}.~~~~~~~~~~~~~~~~~~~~~~  \begin{wraptable}{r}{0.45\textwidth}
 \centering
 \footnotesize
\begin{tabular}[h]{ccl}
\hline 
 {\textbf{Words}}& &{\textbf{POS Tags}} \\
\hline 
\<محرز >&$\rightarrow$&
N\_PROP\\
 \<ينتقل > &$\rightarrow$
 &  VERB\\ 
\<الى > &$\rightarrow$& PREP \\
 \<برشلونة > &$\rightarrow$ &N\_PROP\\
 \<مقابل > &$\rightarrow$& NOUN  \\
120  &$\rightarrow$& NUM \\
\<مليون > &$\rightarrow$& NOUN   \\
 \<دولار> &$\rightarrow$&
 NOUN \\
\hline
\end{tabular} 
\caption{\small POS tags of our example} \label{tab:pos}
\end{wraptable}

\paragraph{Step 2: POS and Token Selection.}   
In this step, tokens corresponding to one or  more POS tags must be chosen for substitution. For our illustrative example, we will select and substitute only the \textit{proper noun} and \textit{digit} tokens. The sentence has two proper nouns, \<برشلونة> and \<محرز>  and one digit ($120$).  
\normalsize\paragraph{Step 3: Sentence Manipulation.} If we select only the noun proper:  \small \<برشلونة>  \normalsize (Barcelona), we can retrieve the $5$-closest words from AraNewsEmb. In this case, we obtain:\small \<مدريد> \normalsize(Madrid),	\small\<ميلان>	 \normalsize(Milan),	\small \<باريس>	 \normalsize (Paris),		\small \<فالنيسيا>	 \normalsize(Valencia), and \small	\<مانشستر>	 \normalsize(Manchester). Indeed, we can generate $5$ fake sentences from the original sentence. However, if we select two proper nouns   \small  \<برشلونة>, \<محرز> \normalsize and the digit token $120$,  we can generate  $75$ ($3*5*5$) manipulated sentences from the single human sentence. Both scenarios are presented in Table~\ref{tab:Fake_sent}.

\begin{table}[H]
\centering
\begin{adjustbox}{width=13cm}

\renewcommand{\arraystretch}{0.8}
{

\scriptsize
\begin{tabular}{c||r}

\toprule
 \textbf{Subs. with $5$-closest of 	\colorbox{blue!10}{\<برشلونة>} }&  \textbf{Subs. with  $5$-closest of  \colorbox{blue!10}{\<برشلونة>},	\colorbox{orange!30}{\< محرز >}	 ~~and 	\colorbox{red!30}{120}}  \\ \toprule

\<دولار> 	\<مليون> 	120	\<مقابل> 	\colorbox{blue!10}{\<مدريد>}	\<الى> 	\<ينتقل> 	\<محرز> 	&

\<دولار> 	\<مليون> 	\colorbox{red!20}{350}	\<مقابل> 	\colorbox{blue!10}{\<ليدز>}	\<الى> 	\<ينتقل> 	\colorbox{orange!30}{\<صلاح>}	\\

\<دولار> 	\<مليون> 	120	\<مقابل> 	\colorbox{blue!10}{\<باريس>}	\<الى> 	\<ينتقل> 	\<محرز> 	& 

\<دولار> 	\<مليون> 	\colorbox{red!20}{450}	\<مقابل> 	\colorbox{blue!10}{\<مدريد>}	\<الى> 	\<ينتقل> 	\colorbox{orange!30}{\<ميسي>}	\\

\<دولار> 	\<مليون> 	120	\<مقابل> 	\colorbox{blue!10}{\<فالنسيا>}	\<الى> 	\<ينتقل> 	\<محرز> 	& 

\<دولار> 	\<مليون> 	\colorbox{red!20}{155}	\<مقابل> 	\colorbox{blue!10}{\<باريس>}	\<الى> 	\<ينتقل> 	\colorbox{orange!30}{\<رونالدو>}	\\

\<دولار> 	\<مليون> 	120	\<مقابل> 	\colorbox{blue!10}{\<ميلان>}	\<الى> 	\<ينتقل> 	\<محرز> 	& 

\<دولار> 	\<مليون> 	\colorbox{red!20}{280}	\<مقابل> 	\colorbox{blue!10}{\<فالنسيا>}	\<الى> 	\<ينتقل> 	\colorbox{orange!30}{\<ماني>}	\\

\<دولار> 	\<مليون> 	120	\<مقابل> 	\colorbox{blue!10}{\<مانشستر>}	\<الى> 	\<ينتقل> 	\<محرز> &   

\<دولار> 	\<مليون> 	\colorbox{red!20}{70}	\<مقابل> 	\colorbox{blue!10}{\<مرسيليا>}	\<الى> 	\<ينتقل> 	\colorbox{orange!30}{\<اغويرو>}



\\


\toprule

\end{tabular}
} \end{adjustbox}

\caption{\small Illustrative output example from our text manipulation method. Given a sentence and a target POS tag, we substitute the word corresponding to the POS tag with the word closest to it (based on cosine similarity) in the AraNewsEmb model. \textbf{Left:} Substitution of word\ \footnotesize \<برشلونة>\  \small(\textit{Barcelona}) with its 5-closets words. \textbf{Right:} Substitution of \footnotesize  \<برشلونة>,   \<محرز>  \small~and \textit{120} (\textit{Barcelona, Mehrez} [name of a soccer player], and 120) each with 5-closest words.}\label{tab:Fake_sent}
\end{table} 



\setcode{utf8}

\normalsize

\section{Human Annotation Study}\label{sec:annot}

\subsection{Annotation Data}
We perform a human annotation study in order to identify (1) the ability of humans to detect machine manipulated text using our method, and (2) the extent to which text identified as machine manipulated can be \textit{fake}. For this purpose, we  randomly select $300$ samples from the ATB development set (see~Table~\ref{tab:data_splits}), among which $145$ sentences are from the original ATB sentences and the rest (i.e., $155$ samples) are machine manipulated. 

\subsection{Annotation Procedures}

\begin{wraptable}{r}{0.51\textwidth}
\begin{adjustbox}{width=8cm}

\renewcommand{\arraystretch}{1}
{
\begin{tabular}{llrrrc}
\toprule
                         &            &             & \multicolumn{2}{c}{\small \textbf{Annotators Agreement (\%)}} &         \\ \cline{3-6}
                         &            & \small \textbf{\#Sent. }& \small \textbf{Hum/Mach }    & \small \small \textbf{True/Fake }   & \small \textbf{\%Fake}  \\
 \toprule
\textbf{\small  Hum }                &            & $145$         & $97.93$                   & N/A             &    N/A       \\ \hline
\multirow{7}{*}{\textbf{\small  Mach}} & \textbf{\small  ADJ} &  $27$	 & $96.30$ & 	$74.07$ & 	$48.15$   \\
 &  \textbf{\small  ADJ\_COMP}   &  $24$ & 	$100$ & 	$91.67$ & 	$58.33$  \\
& \textbf{\small ADJ\_NUM }    & $26$ & 	$76.92$ & 	$73.08$ & 	$78.85$\\
& \textbf{\small  NEG\_PART}    & $32$ & 	 $87.50$ & 	$90.63$ & 	$76.56$  \\
 &\small  \textbf{\small  N\_NUM}     & $19$ & 	$100$ & 	$73.68$ & 	$76.32$  \\
& \textbf{\small  N\_PROP }   & $27$ & 	$92.59$ & 	$74.07$ & 	$83.33$ \\
 \cdashline{2-6}
& \textbf{Overall }      & $155$ & 	$94.67$ & 	$80$ & 	$70.32$  \\
\toprule
\end{tabular}
}\end{adjustbox}
\caption{\small Percentages of inter-annotator agreement on a random sample of 300 sentences (original and manipulated).} \label{kappa}
\end{wraptable}
  For annotation, we follow two stages: The first stage is for \textbf{manipulated text detection}. We shuffle the samples and ask the annotators to label each sentence as either original/produced by humans (\textit{human}) or generated by machine (\textit{machine}). The second stage is for detecting \textbf{veracity of manipulated text}. This stage is applied only on the $155$ machine manipulated sentences generated from ATB. Note that here we provide annotators with a sentence \textit{pair} including the machine generated sentence itself and its human counterpart (original sentence in ATB). Annotators are then asked to compare the manipulated sentence to its original and assign the label \textit{fake} if the manipulated sentence differs in meaningful ways (e.g., provides contradictory information) from the original, but a \textit{true} label otherwise. That is, a \textit{true} tag is assigned if difference between the sentence pair is only grammatical such as cases where the machine sentence is a paraphrase. Each sample is annotated by two experts, both of whom is native speakers of Arabic with a Ph.D. degree. Inter-annotator agreement in term of Kappa ($\kappa$) scores is $79.46\%$ for \textit{human} vs. \textit{machine} and $81.07\%$ for \textit{fake} vs. \textit{true}.  \noindent As shown in Table~\ref{kappa}, the substitution  of tokens with the POS tags ADJ\_NUM, NEG\_PART, N\_NUM, and N\_PROP changes between $76.32$\% and $83.33$\% of sentence veracity. Meanwhile, changing tokens whose POS tags are ADJ or ADJ\_COMP changes the veracity of the sentence less than $50\%$ of the time. The reason is that the selected $k-$closets tokens in the second scenario is more or less of a paraphrase. Table~\ref{tab:disagremment} provides examples where annotators disagree on either or both tasks, and Table~\ref{tab:annotation} illustrates cases where annotators agree. \\  

\begin{table}[h]
\begin{adjustbox}{width=\textwidth}
\renewcommand{\arraystretch}{1.3}
{

\begin{tabular}{ccccclr}

\toprule

  \multicolumn{2}{c}{\textbf{Annotator 2}}  & \multicolumn{2}{c}{\textbf{Annotator 1 }} &  \multirow{2}{*}{\textbf{ POS}}      &  \multirow{2}{*}{\textbf{Gold}}    &      \multirow{2}{*}{\textbf{Sentence}~~~~~~~~~~~~~~~~~~~~~~~~~~~~~~~~~~~~~~}  \\  \cline{1-4}

 \textbf{\small T/F }  &    \textbf{\small M/H}&   \textbf{\small T/F }     &      \textbf{\small M/H}    &      &   &   \\

\toprule



\multirow{2}{*}{{\small True}} & \multirow{2}{*}{{\small {Mach} } }  &  \multirow{2}{*}{{\small Fake }}   &  \multirow{2}{*}{{\small {Mach} } } &  \multirow{2}{*}{{\small N\_PROP }} & Hum  &

\<ذخائر القديسه تريز \textbf{المراهق} يسوع في القبيات واحتفالات دينيه تكريما لها حتي الخميس	>      \\
 
   & & & & & Mach & 
\<ذخائر القديسه تريز \textbf{الطفل} يسوع في القبيات واحتفالات دينيه تكريما لها حتي الخميس> \\\hline

\multirow{2}{*}{{\small True}} & \multirow{2}{*}{{\small {Mach} } }  &  \multirow{2}{*}{{\small Fake }}   &  \multirow{2}{*}{{\small {Hum} } } &  \multirow{2}{*}{{\small N\_PROP }} & Hum  &

\<هذه المؤامره تستهدف احباط العمليه السياسيه وارجاع \textbf{العراق} الي اه حكم اه العصابه البعثيه  >      \\
 
   & & & & & Mach & 
\<هذه المؤامره تستهدف احباط العمليه السياسيه وارجاع \textbf{الاردن} الي اه حكم اه العصابه البعثيه >     \\ \hline

\multirow{2}{*}{{\small Fake }} & \multirow{2}{*}{{\small {Hum} } }  &  \multirow{2}{*}{{\small True }}   &  \multirow{2}{*}{{\small {Mach} } } &  \multirow{2}{*}{{\small ADJ }} & Hum  &

\<وتابع ان التوجه الاعلامي الجديد  \textbf{مرفوض} والحكومه تطلب منا المستحيل>      \\
 
   & & & & & Mach & 
\<وتابع ان التوجه الاعلامي الجديد  \textbf{مضلل}  والحكومه تطلب منا المستحيل > \\ \hline

\multirow{2}{*}{{\small Fake }} & \multirow{2}{*}{{\small {Hum} } }  &  \multirow{2}{*}{{\small Fake }}   &  \multirow{2}{*}{{\small {Mach} } } &  \multirow{2}{*}{{\small NEG\_PART }} & Hum  &

\<واوضح ان مقعدين \textbf{لم} يتم البت بهما بعد في الاسكندريه وان ذلك عائد الي قرار قضائي>    \\
 
   & & & & & Mach & 
   
\<واوضح ان مقعدين يتم البت بهما بعد في الاسكندريه  وان ذلك عائد الي قرار قضائي>    \\ \hline

\multirow{2}{*}{{\small Fake }} & \multirow{2}{*}{{\small {Hum} } }  &  \multirow{2}{*}{{\small True }}   &  \multirow{2}{*}{{\small {Mach} } } &  \multirow{2}{*}{{\small ADJ\_NUM }} & Hum  &

\<ابيكم تعطوني سعر هالقطعه هذي ووين تركب حقت اي بندق \textbf{ثانية} من نفس الصناعه  >   \\
 
   & & & & & Mach & 
   
\<ابيكم تعطوني سعر هالقطعه هذي ووين تركب حقت اي بندق \textbf{رابع} من نفس الصناعه >   \\

\toprule

\end{tabular} }
\end{adjustbox}
\caption{\small Examples of disagreement between annotators on either one or the two tasks.} \label{tab:disagremment}
\end{table}

\begin{table}[h]
\begin{adjustbox}{width=\textwidth}
\renewcommand{\arraystretch}{1.4}
{
\footnotesize
\begin{tabular}{rlccc}

\toprule
 \multirow{2}{*}{\textbf{Sentence~~~~~~~~~~~~~~~~~~~~~~~~~~~~~~~~~~~~~~~~~~~~~~~~~~~~~~~~~~~~~~~~}}  & \multirow{2}{*}{\textbf{Gold}}    &   \multirow{2}{*}{\textbf{ POS}}      & \multicolumn{2}{c}{\textbf{Task Labels}}  \\  \cline{4-5}
   &          &      & \textbf{\small H/M} & \textbf{T/F}\\
\toprule

\<وصدر بيان عن اتحاد علماء المسلمين في \textbf{العراق} جاء فيه>     & Hum &\multirow{2}{*}{{\small N\_PROP } }      &  \multirow{2}{*}{{\small Hum } }       &    \multirow{2}{*}{{\small Fake }}   \\
\<وصدر بيان عن اتحاد علماء المسلمين في \textbf{الاردن} جاء فيه>     & Mach   &&&   \\\cline{1-5}


\<حياك \textbf{الله} اخي الغالي >    & Hum &     \multirow{2}{*}{{\small N\_PROP }} &  \multirow{2}{*}{{\small Hum } }       &    \multirow{2}{*}{{\small True }}   \\
\<حياك \textbf{الرحمن} اخي الغالي > &  Mach &&& \\\toprule

\<هذه  الصور اعادت الي اذهان مشاهد مقتل نحو واحد \textbf{وعشرين} ألف شخص >   & Hum &\multirow{2}{*}{{\small N\_NUM } }     &  \multirow{2}{*}{{\small Hum } }       &    \multirow{2}{*}{{\small  Fake }}   \\
\<هذه  الصور اعادت الي اذهان مشاهد مقتل نحو واحد \textbf{وثلاثين} ألف شخص >    &Mach &&& \\\cline{1-5}

\<احتل الهولندي كارستن المركز الاول في المرحله  الثامنة من دوره فرنسا ال \textbf{81} للدراجات >   & Hum &      \multirow{2}{*}{{\small N\_NUM } }&  \multirow{2}{*}{{\small Hum } }       &    \multirow{2}{*}{{\small  Fake }}   \\
\<احتل الهولندي كارستن المركز الاول في المرحله  الثامنة من دوره فرنسا ال \textbf{89} للدراجات >    &    Mach &&& \\\cline{1-5}


\toprule

\<  يلف الغموض \textbf{العديد} من المشكلات في اداء الجيش الاسرائيلي>   &  Hum &\multirow{2}{*}{{\small ADJ } }      &  \multirow{2}{*}{{\small Hum } }       &    \multirow{2}{*}{{\small  True }}   \\
\< يلف الغموض \textbf{الكثير} من المشكلات في اداء الجيش الاسرائيلي > &  Mach &&& \\ \cline{1-5}

\<احترم نفسك \textbf{احسن} لك والا ساشن حمله لمقاطعه مدونتك ابداها بتوقيعات الزملاء السعوديين >  & Hum & \multirow{2}{*}{{\small ADJ } }    &  \multirow{2}{*}{{\small Hum } }       &    \multirow{2}{*}{{\small  True }}   \\
\<احترم نفسك \textbf{افضل} لك والا ساشن حمله لمقاطعه مدونتك ابداها بتوقيعات الزملاء السعوديين >   & Mach &&& \\\cline{1-5}


\toprule
\< الصابونجي : الهجره المسيحيه \textbf{لا} تتصل  بموضوع ديني>  &  Hum &\multirow{2}{*}{{\small NEG\_Part } }      &  \multirow{2}{*}{{\small Hum } }       &    \multirow{2}{*}{{\small  Fake }}   \\
\<الصابونجي : الهجره المسيحيه تتصل بموضوع ديني> &  Mach &&& \\\cline{1-5}

\<واضاف  كورماك ان واشنطن تفكر في الخضوع في المستقبل في حال \textbf{لم} تنجح الضغوط السياسيه>  & Hum &      \multirow{2}{*}{{\small NEG\_Part }}&  \multirow{2}{*}{{\small Hum } }       &    \multirow{2}{*}{{\small  True }}   \\
\<واضاف  كورماك ان واشنطن تفكر في الخضوع في المستقبل في حال  تنجح الضغوط السياسيه> & Mach &&& \\\cline{1-5}


\toprule
\end{tabular} }
\end{adjustbox}
\caption{\small{Example labels from one annotator on a sample of our data.}} \label{tab:annotation}
\end{table}

\normalsize




\section{Manipulated Text and~Fake News~Detection} \label{sec:dete}
\subsection{Manipulated Text Detection (MTD)}
\label{sub-sec:MTD}
\noindent\textbf{Approach.}  We  use the ATB$^\textbf{+}$  and AraNews$^\textbf{+}$  datasets  for  training  deep  learning  models  for detecting manipulated text. From each of these datasets, we select $61$K human and $61$K machine manipulated sentences (total $\sim 122$K) and split them into 80\% training (TRAIN), 10\% development (DEV), and 10\% test (TEST) as shown in Table~\ref{tab:selec-data}.

\begin{table}[H]
\begin{adjustbox}{width=\textwidth}

\renewcommand{\arraystretch}{1.4}
{\footnotesize
\begin{tabular}{lrrrrrrrr}

\toprule
 \multirow{2}{*}{\textbf{ \# Split} }  & \textbf{ Human}          & \multicolumn{7}{c}{\textbf{ Machine Manipulated }}                          \\ \cline{2-9}
& \textbf{\small \# Sent.} & \textbf{ \footnotesize ADJ}   & \textbf{\footnotesize ADJ\_COMP} & \textbf{\footnotesize ADJ\_NUM} & \textbf{\footnotesize N\_NUM} & \textbf{\footnotesize N\_PROP} & \textbf{\footnotesize NEG\_PART} & \textbf{\footnotesize Total}  \\\toprule
 \textbf{TRAIN} &$ 48,727        $ & $ 9,600 $ & $ 4,513     $ & $ 5,752    $ & $ 9,600     $ & $ 9,600      $ & $ 9,600     $ & $ 48,665 $\\   
\textbf{DEV}    & $ 6,573          $ & $ 1,300 $ & $ 638       $ & $ 844      $ & $ 1,300     $ & $ 1,300      $ & $ 1,300     $ & $ 6,682$  \\
\textbf{TEST}   & $ 5,895          $ & $ 1,200 $ & $ 592       $ & $ 665      $ & $ 1,200     $ & $ 1,200      $ & $ 1,200     $ & $ 6,057$  \\\toprule

\end{tabular}}
\end{adjustbox}

\caption{\small  The TRAIN, DEV, and TEST splits form ATB$^\textbf{+}$ (with a similar split from AraNews$^\textbf{+}$) for developing our manipulated news detection models. The same amount of data from the different POS categories is extracted from each of the two datasets. }  \label{tab:selec-data}
\end{table}

\noindent\textbf{Models.} 
For the purpose of training our  manipulated text detectors,  we exploit  4  large pre-trained masked language models (MLM):   mBERT~\cite{devlin2018bert},  AraBERT~\cite{antoun2020arabert},  XLM-R\textsubscript{Base}, and XLM-R\textsubscript{Large}~\cite{conneau-etal-2020-unsupervised}.~\footnote{Each of the mBERT, AraBERT, and XLM-R\textsubscript{Base} models has $12$ layers each with $12$ attention heads, and $768$ hidden units. The XLM-R\textsubscript{Large} model has $24$ layers each with $16$ attention heads, and $1,024$ hidden units.}\\

\noindent\textbf{Training  Data \&  Hyper-Parameters.} We fine-tuned  all these models  on the TRAIN split of (1)  ATB$^+$, and (2) AraNews$^+$, independently. For each model, we run for $25$ epochs with a batch size of $32$, maximum sequence length of $128$ tokens, and a learning rate of $1e^{-5}$.  \\

\noindent\textbf{Evaluation Data.} We evaluate each of the two models on its respective DEV and TEST splits (i.e., from either ATB$^+$ or AraNews$^+$). Although the data in the two classes are reasonably balanced, we use \textit{both} accuracy and macro $F_{1}$ for evaluation. Table~\ref{tab:res-fake} shows the results on the two datasets. \\


\begin{wraptable}{r}{0.51\textwidth}
 	
\small
\begin{adjustbox}{width=8cm}
\renewcommand{\arraystretch}{1.3}{
\centering

\begin{tabular}{lllccc}
\toprule
\multirow{2}{*}{\textbf{Data}}  & \multirow{2}{*}{\textbf{ Models}} & \multicolumn{2}{c}{\textbf{Dev}}     & \multicolumn{2}{c}{\textbf{Test}}       \\  \cline{3-6} 
      &      &   \textbf{Acc.} & \textbf{ F1}         & \textbf{Acc.}    & \textbf{F1}        \\\toprule
 \multirow{3}{*}{\small \colorbox{blue!10}{\textbf{ATB}$^+$}}    & mBERT     &  $77.16$ & $77.08$ & $77.42$ & $77.36$ \\ 
    &     XLM-R\textsubscript{Base}  &$81.72$ & $81.72$ & \bf 83.22 & $\textbf{83.20}$ 

    \\ 
    &      XLM-R\textsubscript{Large}           &$82.41$ &  $82.38$ & $81.38$ & $81.36$ \\ 
   &      AraBERT        &  $\mathbf{83.19}$ & $\mathbf{83.17}$ &  $82.63$ & $82.62$ \\  \hline

   \multirow{3}{*}{\colorbox{blue!10}{\textbf{AraNews}$^+$}}   & mBERT          & $79.39$ & $79.38$ & $83.51$ & $83.52$ \\
  &     XLM-R\textsubscript{Base}            &       $82.77$&      $82.56$  &         $86.09$&         	$86.08$     \\

     &      XLM-R\textsubscript{Large}           &    $82.12$ & $82.10$ & $86.35$ & $86.35$      \\
     &      AraBERT        &  $\mathbf{87.21}$ & $\textbf{87.21}$ & $\mathbf{89.23}$ & $\mathbf{89.25}$ \\\hline
     

\end{tabular}
}\end{adjustbox}

\caption{\small Performance results of our the MTD  models on the dev and test split of ATB$^\textbf{+}$ and AraNews$^\textbf{+}$.}
\label{tab:res-fake}
\end{wraptable}

\noindent\textbf{\textbf{Results \& Discussion.} } As Table~\ref{tab:res-fake} shows, the best performance on ATB$^+$ is at $83.20$ $F_{1}$ (acquired with XLM-R\textsubscript{Base}). For AraNews$^+$, the best model is at $89.25$ $F_{1}$ (acquired with AraBERT). These results show that it is harder to detect manipulated text exploiting ATB$^+$ than that exploiting AraNews$^+$. This could be due to two reasons: (a) ATB$^+$ contains news stories that are diachronically different from the data the language models are trained on, which is less true for the case of AraNews (since the latter dataset is crawled in late 2019 and early while most ATB data were acquired prior to 2004), (b) ATB$^+$ is POS tagged manually, which makes generations based on it less error-prone. \\


\subsection{Fake News Detection (FND)\\}\label{subsec:FND}

\noindent{\textbf{Approach.}} Evaluating on an external human-crafted fake news dataset, we also develop a host of models for detecting fake news. The dataset is developed by~\newcite{khouja2020stance} by sampling a subset of news titles from the Arabic News Texts corpus \cite{chouigui2017ant}, a collection of Arabic news from multiple news media sources in the Middle East. Crowd-sourcing is used to generate true and false claims starting from a news title. ~\newcite{khouja2020stance} asks annotators to modify each news title into a new claim by: (1)  paraphrasing the original title via changing wording and syntax while maintaining the same meaning, thus producing a legit (or \textit{true}) sentence, and (2) modifying the meaning of the original title such that a sentence that contradicts that title is acquired (constituting a false, or \textit{fake}, claim). We refer to this dataset from \cite{khouja2020stance} as \textbf{\texttt{Khouja}}. It comprises $3,072$ \textit{true} sentences  and  $1,475$ \textit{fake} sentences. We now describe our various fake news detection models. \\

\noindent{\textbf{Models.}} As explained, our primary goal is to test how data generated by our methods will fare on the problem of fake news detection, as evaluated on a human-created fake news dataset (i.e., Khouja). For this reason, we only test models reported in this section on the DEV and TEST splits of Khouja. We have the following modeling settings:

\begin{enumerate}
    \item \textbf{Fine-Tuning on Khouja (Baseline).} Here, we fine-tune  all MLMs (i.e., models from Section~\ref{sub-sec:MTD}) on the train split of Khouja.
    \item \textbf{Zero-Shot Detection.} Based on our human annotation study (Section~\ref{sec:annot}), we hypothesize that our machine-manipulated sentences will be closer to the \textit{fake} class than the \textit{true} class in the fake news context. To test this hypothesis, we fine-tune our MLMs \textit{only} on our generated data (and hence naming this setting \textit{zero-shot}, i.e., since we do not train on Khouja TRAIN at all). We have the following configurations pertaining the parts of our data we fine-tune on: (a) ATB$^+$ TRAIN, (b) AraNews$^+$ TRAIN, and (c) double the size the TRAIN  of AraNews$^+$.
    \item \textbf{Data augmentation.}  We augment the Khouja TRAIN split with the 3 training configurations from our data listed in the zero-shot setting above (i.e., a, b, and c), each time fine-tuning on Khouja and one of these 3 splits. 

\end{enumerate}

\noindent\textbf{Evaluation Data \& Hyper-Parameters.} For the current experiments, as explained earlier, we use the original split of~\newcite{khouja2020stance} (i.e., 80\% TRAIN, 10\% DEV, and 10\% for TEST). We evaluate all the FND models on the DEV and TEST splits of Khouja and use the same hyper-parameters as in Section~\ref{sub-sec:MTD}. \\ 

\begin{wraptable}{r}{0.50\textwidth}
 	
\small
\begin{adjustbox}{width=8cm}
\renewcommand{\arraystretch}{1.3}{
\small
\centering
\begin{tabular}{cclcccc}

\toprule
\multirow{2}{*}{\textbf{Setting}} &\multirow{2}{*}{\textbf{TRAIN Split}}  & \multirow{2}{*}{\textbf{ Model}} & \multicolumn{2}{c}{\textbf{DEV}}     & \multicolumn{2}{c}{\textbf{TEST}}       \\  \cline{4-7} 
     &  &      &   \textbf{Acc.} & \textbf{ F1}         & \textbf{Acc.}    & \textbf{F1}        \\\toprule
  \multirow{4}{*}{\rotatebox[origin=c]{90}{\textbf{\colorbox{blue!10}{Baseline}}}}
  
  & \multirow{4}{*}{\small \textbf{KH}}    & mBERT          & $\textbf{73.40}$&			$64.74$&			$70.39$&			$61.93$ \\
   &   &     XLM-R\textsubscript{Base}  & $72.74$&			$64.27$&			$72.15$&			$64.92$ \\
  &    &      XLM-R\textsubscript{Large}           & $71.52$&			$\textbf{65.60}$&			$72.15$&			$\textbf{67.21}$ \\ 
  &    &      AraBERT        & $73.07$&			$67.10$&			$\bf72.59$&			$67.05$ \\ \cline{2-7}
  \hline 
  
   \toprule
\multirow{12}{*}{\rotatebox[origin=c]{90}{\textbf{\colorbox{blue!10}{Zero-Shot}}}}   & \multirow{4}{*}{\small \textbf{(a)}}    & mBERT     & $61.92$	&$48.14$	&$60.96$	&$49.12$ \\ 
   &   &     XLM-R\textsubscript{Base}  &$61.81$	&$47.42$	&$60.53$	&$47.37$ \\
 &    &      XLM-R\textsubscript{Large}           & $\textbf{62.36}$&  $49.52$& $\bf62.28$& $50.28$ \\
  &    &      AraBERT        & $62.03$	&$47.72$	&$61.62$	&$49.27$ \\  \cline{2-7}


     &  \multirow{4}{*}{\small\textbf{ (b)} }   & mBERT           & $53.09$	&$49.12$	&$53.73$	&$50.70$\\

 &  &     XLM-R\textsubscript{Base}  &   $58.28$	&$47.66$	&$57.89$	&$48.59$\\ 
  &    &      XLM-R\textsubscript{Large}             &$58.06$ & $46.99$ &$61.18$ & $\textbf{52.71}$\\

  &    &      AraBERT        &    $54.42$	& $\textbf{49.94}$	&$53.29$	&$50.12$ \\ \cline{2-7}
  
  
  &  \multirow{4}{*}{\small \textbf{(c)} }   & mBERT           &  $55.41$	&$48.87$	&$54.61$	&$49.18$\\

 &  &     XLM-R\textsubscript{Base}  &    $55.85$	&$48.21$	&$56.58$	&$48.77$ \\ 
  &    &      XLM-R\textsubscript{Large}            &$56.62$ & $48.75$ &$57.89$ & $50.33$ \\
  &    &      AraBERT        &    $54.86$	&$48.65$	& $57.24$	& $51.49$ \\ \cline{2-7}
  

  \hline 
      \toprule
    \multirow{12}{*}{\rotatebox[origin=c]{90}{\textbf{\colorbox{blue!10}{Data Augmentation}}}}  &  \multirow{4}{*}{\small \textbf{KH+(a)}}   & mBERT          & $71.96$ & $65.51$ & $68.20$ & $60.72$ \\

 &  &     XLM-R\textsubscript{Base}  & $70.86$ & $62.39$ & $69.96$ & $62.71$ \\ 
 &    &      XLM-R\textsubscript{Large}           &$65.89$ & $61.40$ &$66.67$ & $62.86$  \\ 

  &    &      AraBERT        & $72.63$ & $67.15$ & $70.83$ & $65.38$ \\ \cline{2-7}
  
  
       &  \multirow{4}{*}{\small \textbf{KH+(b)}}   & mBERT          & $70.20$ & $64.68$ & $69.74$ & $64.58$ \\
  &  &     XLM-R\textsubscript{Base}  & $72.52$ & $67.05$ & $72.37$ & $67.40$ \\
  &    &      XLM-R\textsubscript{Large}           &  $\textbf{73.29}$& $65.71$ & $72.37$ & $65.79$\\

   &    &      AraBERT        & $72.96$ & $62.94$ & $73.90$ & $66.44$ \\ \cline{2-7}

   &  \multirow{4}{*}{\small \textbf{ KH+(c)}}   & mBERT          & $69.54$ & $64.79$ & $68.42$ & $64.11$ \\
  &  &     XLM-R\textsubscript{Base}  & $69.65$& $64.65$ & $72.15$ & $66.94$ \\
  &    &      XLM-R\textsubscript{Large}           & $71.85$ & $\textbf{67.15}$ & $\textbf{74.12}$ & $\textbf{70.06}$ \\ 
   &    &      AraBERT        & $70.20$	&$65.38$&	$73.03$	&$69.90$ \\ \cline{2-7}


  \toprule

\end{tabular}
} \end{adjustbox}

\caption{\small Performance results of our the MTD models on the DEV and TEST splits of Khouja. \textbf{KH}: refer to Khouja TRAIN split.  \textbf{(a)} ATB$^+$, \textbf{(b)} AraNews$^+$, and \textbf{(c)} $2$x AraNews$^+$.  }
\label{tab:res}
\end{wraptable}
 
\noindent\textbf{\textbf{Results \& Discussion.} } As Table~\ref{tab:res} shows, best performance when training on \textit{Khouja TRAIN (gold, our baseline)} is $67.21\ F_{1}$ (acquired with XLM-R\textsubscript{Large}). This is already $2.91\%$ points higher than the best system reported by~\newcite{khouja2020stance} ($64.30\ F_{1}$, not shown in Table~\ref{tab:res}).

For our \textit{zero-shot experiments}, our best model is at $52.71\ F_{1}$ when training on AraNews$^+$ base setting (i.e., setting \textbf{a} in Table~\ref{tab:res}, with TRAIN data = $48,655$ sentences). This result shows that use of data generated by our method is effective on the fake news detection task, even without access to gold training data. In particular, the $52.71\ F_{1}$ we acquire is higher than the baseline majority class in ~\newcite{khouja2020stance} ($40.20\ F_{1}$) and close to their $53.10\ F_{1}$ character-level LSTM model trained on gold data.

Our \textit{data augmentation experiments} show that using double-sized generated data from AraNews (Train= $97,310$ sentences, our setting \textbf{c}) is most effective and results in $70.06\ F_{1}$. \textbf{\textit{This is the best model we report in this paper. It is $\sim 2.85\ F_{1}$ higher than our own baseline, and $5.76\ F_{1}$ better than ~\newcite{khouja2020stance}'s best model. Overall, our results clearly demonstrate the positive impact of our manipulated data on the fake news detection task, thereby lending value to our novel machine generation method.}} \\

\vspace{-5mm}




\section{Conclusion}\label{sec:conc}
We presented a novel, simple method for automatic generation of Arabic manipulated text for the news domain. To enable off-the-shelf use with our method, we also collected and released a new POS-tagged Arabic news dataset. Exploiting our dataset, we developed and released the first Arabic model for detecting manipulated news text. We performed a human annotation study shedding light on the impact of our text manipulation approach on news veracity. Finally, we leveraged our generated data for augmenting gold fake news data from an external source and report a new SOTA on the task of fake news detection. 

In the future, we plan to explore applying our method to languages other than Arabic. This should be straightforward, since the method itself is language-agnostic and only needs a POS tagger and a dataset from a given language. We also plan to investigate more sophisticated text manipulation methods, exploiting data from different domains. We will also study the impact of these methods on detection of machine generated text as well as fake news detection. 


 \section*{Acknowledgements}
MAM gratefully acknowledges support from the Natural Sciences and Engineering Research Council of Canada, the Social Sciences Research Council of Canada, Compute Canada (\url{www.computecanada.ca}), and UBC ARC–Sockeye (\url{https://doi.org/10.14288/SOCKEYE}).

\newpage

\bibliographystyle{coling}
\bibliography{coling2020}

\newpage
\appendix
\appendixpage
\addappheadtotoc
\setcounter{table}{0}  \renewcommand{\thetable}{\Alph{section}\arabic{table}}


\section{AraNews Data }\label{sec:appendix}

\subsection{AraNews: Country, Domain, and Statistics   } 
\label{subsec:appendix-AraNews-stat}

\begin{table}[H]

    \centering
    \begin{adjustbox}{width=\textwidth}
    \renewcommand{\arraystretch}{0.35}{

    \begin{tabular}{lcrlrr}
      
   \toprule
    \textbf{Country} & \textbf{\# Newspaper} & \multicolumn{2}{c}{\textbf{~~~~~~Newspaper Name}}   & \textbf{\#News/Newspaper} & \textbf{\#News/Country}  \\ 
\toprule    
   
&	&	\tiny{\<الشارع 20> }  & Rue20&	\small  $36,556$  &	\\
&	&	\tiny{\<خبر المغرب> }&	Khabarmaroc&	\small  $2,196$ &	\\
&	&	\tiny{\<يا بلادي> }&	Yabiladi&	\small  $28,760$ &	\\
Morocco &	7 &	\tiny{\<البيضاوي> }&	Albidaoui&	$14,019$&	\small  $178,911$   \\
&	&	\tiny{\<الأسد> }&	Assdae&	\small  $18,600$  &	\\
&	&	\tiny{\<الصباح> }&	Assabah&	\small  $68,564$ &	\\
&	&	\tiny{\<الأخبار> }&	Alakhbarpressma&	$1,021$ &	\\  \hline
&	&	\tiny{\<الشروق > }&	Echoroukonline&	\small  $187,936$  &	\\
&	&	\tiny{\<الخبر> }&	Elkhabar&	\small  $121,441$  &	\\
Algeria &	6 &	\tiny{\<الشعب> }&	Ech chaab&	$147,960$&	$520,162$    \\
&	&	\tiny{\<المساء> }&	el-massa&	\small  $59,917$ &	\\
&	&	\tiny{\<الجديد اليومي> }&	Eljadidelyawmi&	\small  $2,556$ &	\\
&	&	\tiny{\<الامة > }&	Alomah&	\small  $352$&	\\ \hline
&	&	\tiny{\<الجريدة> }&	Aljaridah&	\small  $44,354$  &	\\
&	&	\tiny{\<الصريح> }&	Assarih&	\small  $99,468$  &	\\
Tunisia&	5&	\tiny{\<المغرب> }&		Lemaghreb	&   $76,550$  &  $451,278 $ \\
&	&	\tiny{\<حقائق اونلاين> }&	Hakaekonline&	\small  $128,553$&    \\
&	&	\tiny{\<الشروق> }&	Alchourouk	& $102,353$&     \\\hline

&	&	\tiny{\<اليوم> }& 	 Elyom	&	  $22,993$  &	  \\
&	&	\tiny{\<الأهالي> }&	Alahalygate &		$25,235$ & \\
Egypt &	5&	\tiny{\<طريق الاخبار> }&		Akhbarway& 	$80,561$&   $3,021,352$    \\	
&	&	\tiny{\<صوت الامة> } &	Soutalomma	& $133,128$&     \\	
&	&	\tiny{\<اليوم 7> }&	Youm7 &		$2,759,435$  &   \\\hline

&	&	\tiny{\<أنحاء> }&	An7a	&   $70,985$ &   \\ 
&	&	\tiny{\<الرياض> }&		Alriyadh	& $212,666$&     \\	 
Saudi&	5&	\tiny{\<أم القرى> }&	Uqngovsa &		$20,994$ & $304,899$ \\
&	&	\tiny{\<الحدث> }&	Alhadath	&	$220$  & \\
&	&	\tiny{\<الجزيرة> }&	Aljazeera &		$34$ &   \\\hline

&	&	\tiny{\<صدى الشام> }&	Sadaalshaamnet	&   $12,994$ &     \\
Syria &	3 &	\tiny{\<الوطن> }&		Alwatansy	& $104,68$& $47,058$    \\	
&	&	\tiny{\<الأيام السورية> }& Ayyamsyrianet&	\small  $23,578$ &  \\ \hline

&	&	\tiny{\<السوداني نيوز> }&			Alsudaninews	&   $11,153$ &   \\	
Sudan &	3 &	\tiny{\<السودان اليوم> }&	Alsudanalyoum	& $10,1924$ & $113,121$   \\	
&	&	\tiny{\<ألوان السودانية> }&	 Alwandaily&	\small  $44$ &  \\ \hline

&	&	\tiny{\<الشارع نيوز> }&	Alsharaeanews&	$1,261$ &  \\	
Yemen&	3&	\tiny{\<الصمود> }&	Alsomoud&	$94,86$&	\small  $83,802$\\
&	&	\tiny{\<الثورة> }&	Althawrah&	$73,055$ & \\ \hline

&	&	\tiny{\<بيروت تايمز> }&	Beiruttimes&	$9,629$  &	\\
USA&	2  &	\tiny{\<صدى الوطن> }&	Sadaalwatan&	$11,091$ & $99,080$\\	
&	&	\tiny{\<وطن سرب> }&	Watanserb&	$78,360$  & \\  \hline

UK&	2 &	\tiny{\<ميدل ايست اونلاين> }&	Middleeastonline&	$295,190$ & $295,566$  \\	
&	&	\tiny{\<بي بي سي> }&	BBC&	$376$  & \\ \hline

UAE &	2 &	\tiny{\<الأيام > }&	Alayam&	$5471$&	\small  $63897$\\
&	&	\tiny{\<البيان> }&	Elbyan&	$58426$  & \\ \hline	
Bahrian &	1 &	\tiny{\<البحرين> }&	Bahrian&	\small  $7,612$&	\small  $7,612$ \\ \hline
Iraq &	1 &	\tiny{\<الزمان> }&	Azzaman&	$120,311$&	\small  $120,311$ \\ \hline
Kuwait&	1&	\tiny{\<صحيفة الوسط> }&	Alwasat&	$31,354$ &	\small  $31,354$\\ \hline
Jordan&	1 &	\tiny{\<الدستور> }&	Addustour&	$689,444$ &	\small  $689,444$ \\ \hline
Lebanon&	1  &	\tiny{\<أخبار الأرز> }&	Cedarnews&	$42,388$ &	\small  $42,388$\\ \hline
Palestine&	1 &	\tiny{\<عرب 48> }&	Arab48&	$35,286$&	\small  $35,286$\\

\toprule
    \end{tabular} }
    \end{adjustbox}
    \caption{\small Descriptive statistics of  our ArNews dataset.}
    \label{tab:araNews}
\end{table}
\subsection{AraNews: Domain Normalization }\label{App:norma}
 \begin{table}[ht]

\footnotesize
\begin{center}
\begin{tabular}{rrrl}
\toprule
 {\textbf{Sub-Categories~~~~~~~~~~~~~~~~~~~~~~~~~~~~~~~~~~~~ ~~~~~~~~~~~~~}}& &  \multicolumn{2}{c}{\textbf{Category~~~~~~~~}} \\
\toprule
\footnotesize{\<الدين والحياة>, \<اسلاميات>, \<الاسلامي> ,\<ثقافة قرأنية>} &$\rightarrow$& \footnotesize{\<الدين>} & \small  Religion \\
\footnotesize{\<الصباح التربوي>, 	\<تربية وتعليم>, 	\<تربية>, 	\<التربية و التعليم>}  &$\rightarrow$ & \footnotesize{\<تعليم>} & \small Education  \\
\footnotesize{\<ثقافي>, 	\<الثقافة>, 	\<ثقافة وفنون>, 	\<منوعات و فنون >, 	\<فن وثقافة>, 	\<الثقافة و الفن>, 	\<ثقافية>}  &$\rightarrow$& \footnotesize{\<ثقافة>} & \small Culture\\

\footnotesize{\<اخبار التكنولوجيا>, 	\<علوم >, 	\<علوم وتك >, 	\<علوم تكنولوجية >, 	\<تكنولوجيا >, 		\<علوم وتكنولوجيا>} &$\rightarrow$&   \footnotesize{\<تكنولوجيا>} &  \small Technology\\

\footnotesize{\<اقتصاد وبورصة>, 	\<اسواق >, 	\<الاخبار الاقتصادية >, 	\<اقتصاد وسياحة    >, 	\<أخبار الاقتصاد >, 		\<مال و اعمال>}  &$\rightarrow$ & \footnotesize{\<اقتصاد>} & \small Economy  \\
\footnotesize{\<مجلس الوزراء>, 		\<مراسيم ملكية >, 		\<قرارات وزارية>, 	\<مجلس النواب >, 	\<الاحزاب >, 	\<برلمان >, 	\<نقابات >, 	\<سياسة>} &$\rightarrow$& \footnotesize{\<سياسة>} & \small  Politics  \\
\footnotesize{\<مواقف رياضية>, 		\<رياضة عالمية>, 	\<أخبار الرياضة>, 	\<رياضة دولية>, 	\<رياضة وطنية>, 	\<رياضة محلية>, 	\<رياضة  >}    &$\rightarrow$& \footnotesize{\<رياضة>} &  \small Sport  \\
\footnotesize{\<العلم والصحة>, 	\<طبّ و صحّة>, 	\<الصحة>, 	\<فايروس كورونا>, 	\<صحة وطب  >, 	\<أخبار الصحة والطب>, 	\<صحة>} &$\rightarrow$&   \footnotesize{\<صحة>} &  \small Health\\
\toprule
\end{tabular}
\caption{ \small Story sub-categories and main categories to which we map in AraNews. \label{tab:norm}}
\end{center}
    \end{table}
\normalsize

    



\subsection{ATB$^+$ and AraNews$^+$ Data Splits }\label{App:data-man}

\begin{table}[h]
\footnotesize
\begin{adjustbox}{width=\textwidth}
\renewcommand{\arraystretch}{1.5}
{
\begin{tabular}{clrrrrrrr}
\toprule
 \multirow{2}{*}{\textbf{\small Data} }  & \multirow{2}{*}{\textbf{\small\# Split} }  &  \textbf{\small Human}         & \multicolumn{6}{c}{\textbf{\small Machine Manipulated }}                          \\ \cline{3-9}
& & \textbf{\small  \# Sent.} & \textbf{ \footnotesize ADJ}   & \textbf{\footnotesize ADJ\_COMP} & \textbf{\footnotesize ADJ\_NUM} & \textbf{\footnotesize N\_NUM} & \textbf{\footnotesize N\_PROP} & \textbf{\footnotesize NEG\_PART}   \\\toprule

& \textbf{TRAIN} &$ 48.7K$ &  $99.5K$ & $4.5K$ & $5.8K$  & $60.6K$    & $75.8K$ & $43.6K$  \\

\multicolumn{1}{c}{\multirow{-1.4}{*}{\rotatebox[origin=c]{90}{\textbf{ \small \colorbox{blue!10}{ATB$^\textbf{+}$}}}}}   &\textbf{DEV}    & $ 6.6K$ &    $13.4K$                         & $638$       & $844$      & $7.1K$     & $10.6K$     & $5.6K$    \\

& \textbf{TEST}   & $5.9K$                    & $11.9K$                         & $592$       & $665$      & $8.1K$     & $9.5K$      & $5K$    \\
\toprule

& \textbf{TRAIN} & $3.27M$ &  $6.2M$ & 	$251.4K$ & 	$298.5K$ & 	$1.4M$ & 	$2.3M$ & 	$387.6K$  \\
\multicolumn{1}{c}{\multirow{-2.1}{*}{\rotatebox[origin=c]{90}{\textbf{\small \colorbox{blue!10}{AraNews$^\textbf{+}$}}}}}   & \textbf{DEV}    & $ 5.51K$ &  $7.8M$	&$290.6K$	&$293.7K$&	$1.4M$&	$3.6M$&	$704.7K$  \\ 
& \textbf{TEST}   & $ 6.16K$ & $64M$ &	$343.6K$ &	$303.4K$	 &$1.3M$ &	$5.8M$ &	$496.9K$ \\   \toprule

\end{tabular}}
\end{adjustbox}

\caption{\small Data splits and distribution of POS tags in our machine manipulated datasets : ATB$^\textbf{+}$ and AraNews$^\textbf{+}$}. \label{tab:man-data} 
\end{table}









\end{document}